\documentclass[10pt,twocolumn,letterpaper]{article}

\usepackage{emorama}

\usepackage{cuted}
\usepackage{tabularx}
\usepackage{array}
\usepackage{pifont}
\usepackage{simpleicons}

\definecolor{linkblue}{rgb}{0.21,0.49,0.74}
\definecolor{hf-yellow}{HTML}{ffd000}
\usepackage[breaklinks,colorlinks,allcolors=linkblue]{hyperref}
\usepackage[capitalize]{cleveref}
\crefname{section}{Sec.}{Secs.}
\Crefname{section}{Section}{Sections}
\crefname{table}{Tab.}{Tabs.}
\Crefname{table}{Table}{Tables}

\newcommand{\cmark}{\ding{51}}
\newcommand{\xmark}{\ding{55}}

\title{
    EmoRama3D: A Human-Annotated Dataset for \\
    Affective Perception in 3D Environments
}

\author{
    Qiqi Liu$^{1,2}$ \quad
    Shengqi Dang$^{1,3}$ \quad
    Jiaxin Li$^{1,4}$ \quad
    Chenxi Zhang$^{1,5}$ \quad
    Nan Cao$^{1,3}$\thanks{Corresponding author: nan.cao@gmail.com}\\[4pt]
    {\small $^{1}$Shanghai Innovation Institute} \quad
    {\small $^{2}$Fudan University} \quad
    {\small $^{3}$Tongji University}\\
    {\small $^{4}$Westlake University} \quad
    {\small $^{5}$Huazhong University of Science and Technology}\\[3pt]
    {\small
    {\raisebox{-0.12em}{\simpleicon{github}}}
    \href{https://github.com/idvxlab/EmoRama3D}{idvxlab/EmoRama3D}
    \qquad
    {\textcolor{hf-yellow}{\raisebox{-0.12em}{\simpleicon{huggingface}}}}
    \href{https://huggingface.co/datasets/SII-iDVX/EmoRama3D}{SII-iDVX/EmoRama3D}
    }
}

\date{}

\begin{document}

\maketitle

\begin{strip}
    \centering
    \includegraphics[width=0.98\textwidth]{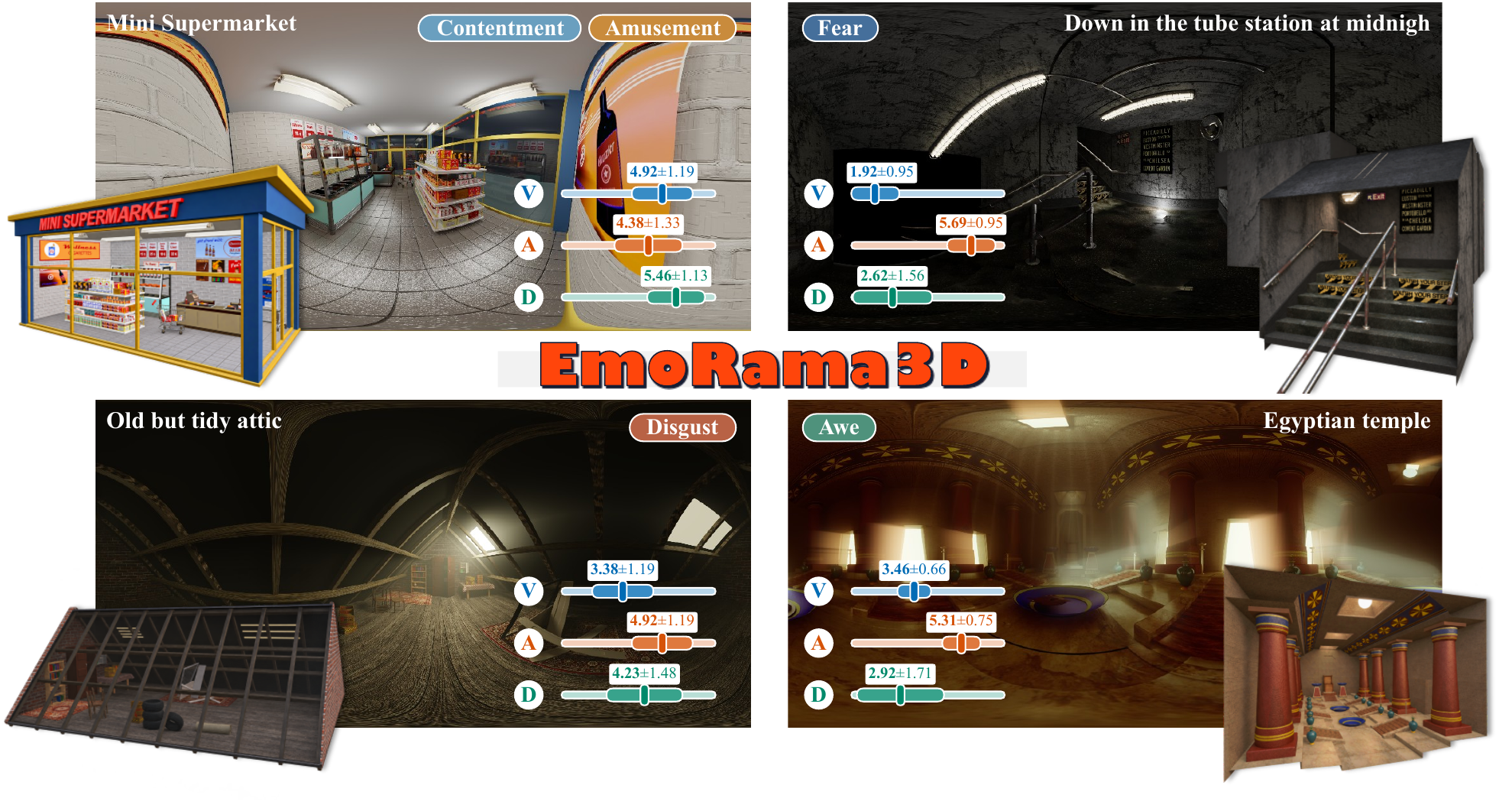}

    \captionsetup{hypcap=false}
    \captionof{figure}{
        We introduce \textbf{EmoRama3D}, a dataset of diverse,
        editable 3D environments with human emotion annotations
        collected through immersive VR. Examples show 360°
        panoramas alongside their source 3D scenes, categorical
        and dimensional emotion annotations. Valence (V),
        arousal (A), and dominance (D) ratings are reported
        as mean ± standard deviation across annotators.
    }
    \label{fig:emorama3d_overview}
    \captionsetup{hypcap=true}
\end{strip}


\begin{abstract}
Understanding human emotional responses to 3D environments
is important for affect-aware spatial design and virtual
environment creation.
However, existing visual affective datasets primarily
focus on images and videos, with limited resources
capturing human emotional responses to immersive
3D environments.
To address this gap, we introduce \textbf{EmoRama3D},
a human-annotated dataset for affective perception
of 3D environments, featuring diverse and editable
3D scenes, immersive VR-based annotations, and
both dimensional and categorical affective
representations (i.e., valence, arousal, and
dominance (VAD) dimensions and Mikels-8
emotion categories).
EmoRama3D comprises 159 editable 3D scenes
and 186 panoramic renderings, annotated by
22 participants through immersive VR.
To construct the dataset, we develop a pipeline
for collecting and rendering 3D scenes,
along with a VR-based annotation system
for capturing human emotional responses.
We further evaluate existing visual models
on predicting human emotional responses
to 3D environments, demonstrating the utility
of EmoRama3D for computational affective
understanding of spatial environments.
\end{abstract}

\begin{table*}[t]
\centering
\caption{
Comparison of visual affective datasets.
VA and VAD denote valence--arousal and
valence--arousal--dominance, respectively.
Physiological signals refer to recorded bodily responses,
such as heart rate and skin conductance.
}
\label{tab:dataset_comparison}
\footnotesize
\setlength{\tabcolsep}{4pt}
\renewcommand{\arraystretch}{1.12}

\resizebox{\textwidth}{!}{%
\begin{tabular}{@{}lcclclll@{}}
\toprule
\textbf{Dataset} &
\textbf{Year} &
\textbf{Venue} &
\textbf{Stimulus} &
\textbf{VR} &
\textbf{Emotion Annotations} &
\textbf{Additional Data} &
\textbf{Dataset Size} \\
\midrule

\multicolumn{8}{@{}l}{
\textit{General Visual Affective Datasets}} \\

ArtEmis~\cite{achlioptas2021artemis}
& 2021
& CVPR
& Art images
& \xmark
& 9 emotions
& Text explanations
& 80K images \\

EmoSet~\cite{yang2023emoset}
& 2023
& ICCV
& Natural/art images
& \xmark
& Mikels-8 emotions
& Visual attributes
& 118K images \\

EmoVerse~\cite{guo2026emoverse}
& 2026
& TAC
& Natural/art images
& \xmark
& Mikels-8 emotions
& Descriptions, embeddings
& 234K images \\

\midrule
\multicolumn{8}{@{}l}{
\textit{Environment Affective Datasets}} \\

EMOIS~\cite{yonekura2026emois}
& 2026
& Preprint
& Built-environment images
& \xmark
& VA
& --
& 1,544 images \\

IDEAS~\cite{deniz2026ideas}
& 2026
& Preprint
& Indoor photographs
& \xmark
& Valence, tense arousal, energetic arousal
& Approach--avoidance
& 1,800 images \\

\midrule
\multicolumn{8}{@{}l}{
\textit{Immersive Affective Datasets}} \\

CEAP-360VR~\cite{xue2023ceap}
& 2021
& TMM
& 360$^\circ$ videos
& \cmark
& VA
& Physiological signals, eye/head tracking
& 8 videos \\

IAVRS~\cite{mancuso2024iavrs}
& 2024
& Sensors
& 360$^\circ$ images
& \cmark
& VAD, 11 emotions
& Physiological signals, eye tracking
& 46 panoramas \\

MERD-360VR~\cite{chen2025merd}
& 2025
& ICMI
& 360$^\circ$ videos
& \cmark
& 6 basic emotions + neutral
& Physiological signals, gaze tracking
& 18 videos \\

MBB-AVR~\cite{fourcade2026mbbavr}
& 2026
& Preprint
& Stereoscopic 360$^\circ$ video
& \cmark
& VA
& Physiological signals, eye/head tracking
& 23-min video \\

\midrule

\textbf{EmoRama3D (Ours)}
& --
& --
& \textbf{360$^\circ$ panoramas}
& \cmark
& \textbf{VAD, Mikels-8 emotions}
& \textbf{Editable 3D scene assets}
& \textbf{186 panoramas (159 scenes)} \\

\bottomrule
\end{tabular}%
}
\end{table*}

\section{Introduction}

Human emotional experiences are shaped not only by
visual content but also by the spatial environments
in which they occur, reflecting the embodied and
situated nature of emotion
~\cite{ferreira2025architecturalreview,chen2026interiors}.
Understanding human affective perception of
3D environments is therefore important for
applications such as immersive interaction,
architectural design, and virtual environment
generation~\cite{ma2025immersivedesign,lin2026library}.
However, developing computational models of
affective perception of 3D environments requires datasets that capture
human emotional responses to diverse
spatial environments.

Existing visual affective datasets primarily
focus on images and videos, such as
ArtEmis~\cite{achlioptas2021artemis},
EmoSet~\cite{yang2023emoset}, and
EmoVerse~\cite{guo2026emoverse},
as summarized in Table~\ref{tab:dataset_comparison}.
More recent datasets, including
EMOIS~\cite{yonekura2026emois} and
IDEAS~\cite{deniz2026ideas}, have extended
affective analysis to photographs of
architectural and indoor spaces.
Meanwhile, virtual reality (VR) has enabled
immersive affective annotation, allowing
participants to experience panoramic
environments and report their emotional
responses~\cite{xue2023ceap,chen2025merd,
mancuso2024iavrs,fourcade2026mbbavr}.
Despite these advances, existing datasets
remain limited in supporting affective
understanding of 3D environments.
First, image-based annotations may not fully
reflect emotional responses under immersive
viewing conditions~\cite{jiang2024interactive}.
Second, existing immersive datasets contain
relatively limited numbers of stimuli,
restricting the diversity of spatial
environments available for computational
modeling.
Third, most immersive datasets do not provide
editable 3D scene assets, limiting their
utility for affect-aware scene manipulation and design.

To bridge this gap, we introduce
\textbf{EmoRama3D}, a human-annotated dataset
for affective perception of 3D environments.
EmoRama3D comprises 159 editable 3D scenes,
186 rendered 360-degree panoramas, with emotional annotations collected from 22 participants.
To construct the dataset, we collect and
curate diverse indoor and outdoor 3D scenes,
and render them into panoramas for
immersive viewing.
We further develop a VR-based annotation
system to collect human emotional responses.
To capture complementary aspects of emotion,
we adopt both dimensional and categorical
representations.
Specifically, the dimensional representation
includes (1) valence, the degree of emotional
pleasantness; (2) arousal, the level of
emotional activation; and (3) dominance,
the degree of perceived control~\cite{russell1980circumplex,bradley1994measuring}.
The categorical representation adopts
Mikels-8, comprising eight emotion categories:
amusement, contentment, awe, excitement,
fear, sadness, disgust, and anger~\cite{mikels2005emotional}.

To demonstrate the utility of EmoRama3D,
we evaluate existing visual models on
predicting human emotional responses
from 3D scene panoramas.
We investigate pretrained visual
representations, model adaptation strategies,
and panoramic viewing configurations.
The results demonstrate the feasibility
of learning human-perceived emotions
from visual scene content, while revealing
remaining challenges in affective
understanding of 3D environments.

In summary, our contributions are:
\begin{itemize}
    \item We introduce \textbf{EmoRama3D},
    a human-annotated dataset comprising
    diverse, editable 3D scenes and their
    rendered panoramas, with emotional
    responses collected through immersive VR.

    \item We provide rich affective annotations
    spanning dimensional VAD ratings and
    categorical Mikels-8 emotions,
    preserving individual variability
    and emotion co-occurrence.

    \item We evaluate
    existing visual models on EmoRama3D
    under different settings,
    demonstrating its utility for
    predicting human emotional responses
    to 3D environments.
\end{itemize}

\section{Related Work}

\subsection{Emotion Representation}
Psychologists often represent emotions using multiple representational paradigms: categorical and dimensional ones. In foundational studies, Ekman~\cite{ekman1992argument} and Mikels~\cite{mikels2005emotional} conceptualize emotions as discrete categories, such as happiness, sadness, and fear. However, discrete emotion labels can hardly cover the diversity and subtle changes of human emotions. The dimensional paradigm quantifies human affect in a continuous space, typically along valence, arousal, and dominance (VAD) dimensions~\cite{russell1980circumplex,bradley1994measuring}. Recent studies have further investigated the relationship between these representations, revealing associations between categorical and dimensional ratings in affective images~\cite{horvat2022investigation}, as well as continuous transitions between emotion categories~\cite{faul2023representation}. In this work, we adopt both representations, using dimensional VAD ratings and categorical Mikels-8 emotion labels to capture complementary aspects of human emotional responses to 3D spaces.

\subsection{Visual Affective Computing}

Visual affective computing aims to understand human emotional responses evoked by visual stimuli. This field has evolved from handcrafted features to deep learning-based features~\cite{zhao2021aica}. Early studies relied on low-level visual descriptors like color, texture and composition~\cite{machajdik2010affective}, whereas later studies integrated higher-level features including roundness, angularity, visual complexity~\cite{lu2017investigation} and particularly high-level semantic concepts~\cite{borth2013large}.

More recently, vision-language pretraining model has opened new directions for visual affective computing. Models such as CLIP~\cite{radford2021learning} and SigLIP~\cite{zhai2023siglip} learn aligned visual-language representations from large-scale image-text pairs, providing rich semantic features for downstream visual understanding~\cite{radford2021learning,zhai2023siglip,tschannen2025siglip2}. Building upon these representations, Bustos et al.~\cite{bustos2023clip} exploited CLIP embeddings for fine-grained visual sentiment classification, while Deng et al.~\cite{deng2024learning} introduced instance-specific prompt tuning to enhance CLIP-based image emotion classification. Beyond these approaches, the general perception and reasoning capabilities of multimodal large language models (MLLMs) have motivated their application to emotion understanding~\cite{cheng2024emotionllama,bhattacharyya2025evaluating}. For instance, Fang et al.~\cite{fang2025catch} proposed an inference framework that combines coarse-to-fine reasoning to improve MLLMs' emotion perception. 
Recent efforts have further explored MLLM-based affective understanding through benchmarks such as EEmo-Bench~\cite{gao2025eemobench}, MultiEmo-Bench~\cite{chen2026multiemo}, and AICA-Bench~\cite{she2026aicabench}, covering image-evoked emotion assessment, multi-annotator emotion distributions, and affective reasoning and generation.

Unlike these works that mainly focus on 2D visual stimuli, EmoRama3D provides 3D scene together with immersive VR-based emotional annotations, enabling computational affective understanding and affect-aware manipulation of 3D spatial environments.

\subsection{Spatial Emotional Perception}

Beyond 2D visual stimuli, 3D spatial environments can evoke human emotional responses through spatial geometry and configuration, enclosure, lighting, colors, materials, and other environmental attributes~\cite{bower2019built,stjean2022review}. Vartanian et al.~\cite{vartanian2013impact} examined aesthetic and affective responses to architectural photographs, while Küller et al.~\cite{kuller2009color} investigated the psychological and physiological effects of colors in full-scale physical rooms. More recently, immersive VR has become an effective medium for architectural affective assessment, as it allows researchers to systematically manipulate spatial properties while maintaining a controlled and immersive experience. Ma et al.~\cite{ma2025immersivedesign} compared VR-based design reviews with 2D visualizations and physical site visits, demonstrating advantages of immersive VR in evaluating spatial properties. Other studies have examined the affective effects of architectural geometry and proportions~\cite{presti2022measuring}, as well as interior colors, materials, and lighting~\cite{lin2026library}. Beyond individual design factors, Vlavianos et al.~\cite{vlavianos2022human} investigated affective and physiological responses during immersive exploration of reconstructed architectural spaces, while Pantilimonescu-Cioanc\u{a} et al.~\cite{pantilimonescu2026facades} employed machine learning to predict emotional responses to parametrically controlled adaptive facades. 

Unlike these studies, which primarily investigate
emotional responses to specific spatial design factors
or architectural settings, our work introduces a dataset
of diverse 3D environments with emotional annotations, supporting computational
modeling of spatial affective perception.

\begin{figure*}[htb]
    \centering
    \includegraphics[width=\linewidth]{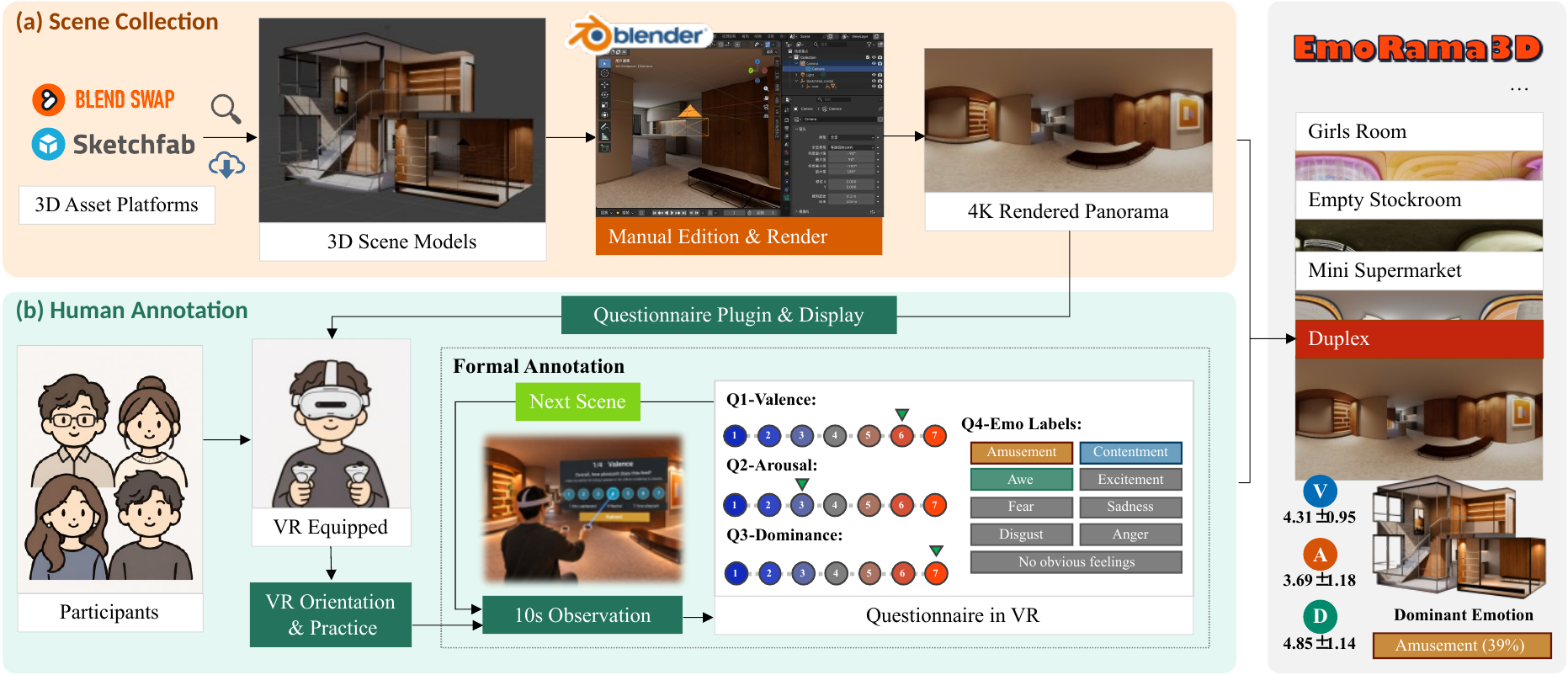}
    \caption{
        Overview of our dataset construction pipeline.
        Existing 3D scene assets are rendered into immersive panoramas
        using Blender and presented to participants through a VR headset.
        Human VAD ratings and Mikels-8 emotion annotations are aggregated
        into panorama-level affective statistics and distributions.
    }
    \label{fig:data_construction}
\end{figure*}
\begin{figure*}[htb]
    \centering
    \includegraphics[width=\textwidth]{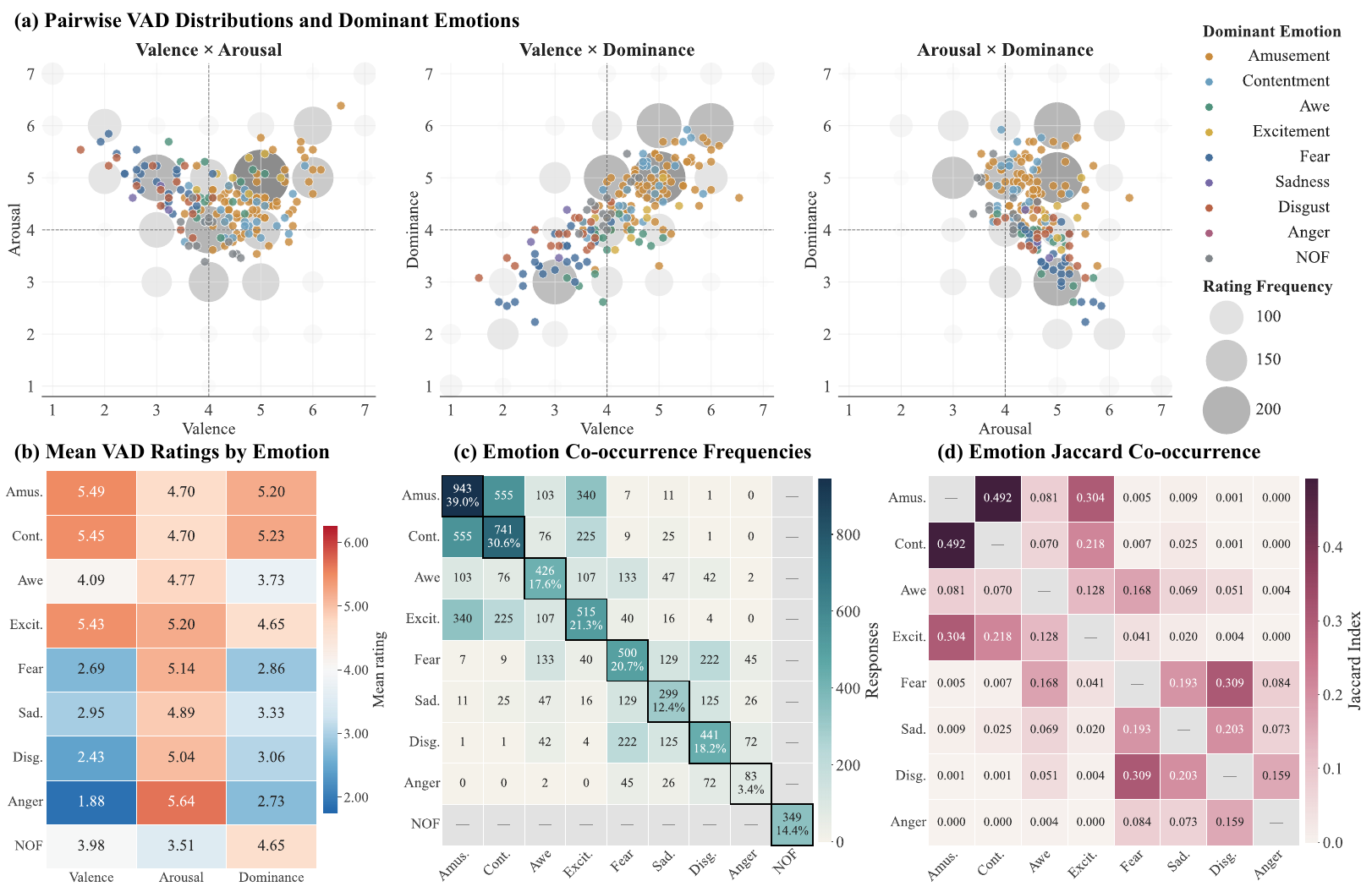}
    \caption{Affective characteristics of EmoRama3D. (a) Pairwise VAD distributions, with point colors indicating dominant emotion categories and bubble sizes representing rating frequencies. (b) Mean VAD ratings conditioned on selected emotion categories. (c) Emotion co-occurrence frequencies. (d) Pairwise emotion associations measured by the Jaccard index.}
    \label{fig:affective_overview}
\end{figure*}

\section{EmoRama3D}
\label{sec:dataset}

EmoRama3D is a human-annotated affective dataset
for understanding human emotional responses to 3D environments.
It comprises 159 editable 3D scenes, 186 rendered
panoramas with affective annotations
collected from 22 participants through immersive VR.

\paragraph{Spatial Content.}
To capture affective responses across diverse spatial
environments, EmoRama3D covers a wide range of indoor
and outdoor spaces with varying spatial layouts,
architectural styles, and visual characteristics.
The dataset provides both explicitly modeled,
editable 3D scenes and their corresponding
rendered panoramas.
The former preserve the underlying geometry and
appearance for scene manipulation and design,
while the latter provide immersive visual stimuli
for affective annotation.

\paragraph{Emotion Representation.}
To capture complementary aspects of human emotion,
we adopt \textit{discrete categories}  and \textit{continuous dimensions} to represent emotions,
as illustrated in Figure~\ref{fig:data_construction}.
The dimensional representation follows
Valence-Arousal-Dominance (VAD)
~\cite{bradley1994measuring}, describing emotional
pleasantness, activation, and perceived control,
respectively.
The categorical representation adopts
Mikels-8~\cite{mikels2005emotional}, comprising
four positive emotions (amusement, contentment,
awe, and excitement) and four negative emotions
(anger, disgust, fear, and sadness), alongside an
additional ``no obvious feeling'' (NOF) option.

\vspace{0.3cm}
We next describe the collection and preparation of
3D scenes (Figure~\ref{fig:data_construction} (a)), the human annotation
procedure (Figure~\ref{fig:data_construction} (b)), and the analysis of the collected annotations.

\subsection{Scene Collection}

To ensure diverse spatial content while preserving
editable 3D representations, we collected 159 3D scenes from BlendSwap~\cite{blendswap} and Sketchfab~\cite{sketchfab_license} under their respective applicable licenses.
The collected assets, provided in \texttt{.blend} or \texttt{.glb} formats, cover a variety of indoor and outdoor environments with diverse spatial layouts and visual characteristics.
All scenes were imported into Blender~\cite{blender} for inspection and preprocessing.
Since assets from different sources may contain missing textures, inconsistent materials, or geometric defects, experienced 3D modelers manually inspected and repaired the scenes to ensure their visual integrity.
Details are provided in the supplementary material.

To provide consistent immersive stimuli for human affective annotation, we further rendered the collected 3D scenes into 360-degree panoramas.
Rather than relying on real-time 3D rendering, which may introduce latency and performance variations across scenes, we adopted an offline
rendering pipeline.
Specifically, we selected fixed viewpoints within the scenes and generated 186 equirectangular panoramas at a resolution of $4096\times2048$.
These panoramas were subsequently presented as surrounding visual environments in VR, allowing participants to explore the scene visually through head rotation while remaining at a fixed viewpoint.

\subsection{Human Annotation}
We conducted an immersive VR annotation study with 22 participants to collect human emotional responses to the 3D environments. The 186 panoramas were evenly divided into three groups, and participants could choose to annotate one, two, or all three groups. Each panorama received annotations from 13 distinct participants, yielding 2,418 annotations in total. Participants were first briefed on the study procedures and provided informed consent. They then received training on the PICO 4 Ultra headset and our VR annotation system to become familiar with scene viewing and in-headset emotion reporting.

Following this familiarization, participants completed a series of annotation trials, each consisting of scene exploration and emotion assessment. In each trial, the system first presented a 360$^{\circ}$ panorama, which participants could freely explore by rotating their heads while remaining at a fixed viewpoint. After 10 seconds of viewing, an in-headset questionnaire appeared. Participants rated valence, arousal, and dominance on seven-point scales and selected one to three Mikels-8 emotion categories, with no obvious feelings (NOF) available as an additional option. Upon submission, the system automatically presented the next panorama after an approximately 7-second interval.
Each annotation session lasted approximately one hour, and participants received appropriate compensation.

\subsection{Dataset Analysis}

We analyze the collected data to characterize the distribution of emotional annotations, co-occurrence patterns among emotion categories, and annotation reliability.

\paragraph{Emotion Distributions.}
We first examine the distribution of dimensional
emotion ratings and their relationship with
categorical responses.
Figure~\ref{fig:affective_overview}(a) shows pairwise
projections of panorama-level VAD ratings, where
colors indicate dominant emotions and bubble sizes
represent rating frequencies.
The ratings span a broad region of VAD space,
with most observations concentrated in the
middle-to-high ranges.
To relate these dimensional ratings to emotion
categories, Figure~\ref{fig:affective_overview}(b)
summarizes the mean VAD ratings associated with
each selected category.
Amusement and contentment have high mean valence
scores (5.49 and 5.45), whereas anger, disgust,
and fear have lower scores (1.88, 2.43, and 2.69).
Anger also combines the highest mean arousal
(5.64) with relatively low dominance (2.73).
These differences illustrate how distinct
categorical responses are reflected in VAD space.

\paragraph{Emotion Co-occurrence.}
We next examine emotion category frequencies and
co-occurrence patterns at the annotation level.
Figure~\ref{fig:affective_overview}(c)
reports selection counts and rates on the diagonal
and pairwise co-occurrence counts off the diagonal,
with NOF summarized separately.
Selection rates represent the percentage of
annotations containing each label.
Amusement (39.0\%) and contentment (30.6\%) are
the most frequently selected categories, whereas
anger (3.4\%) is the least frequent.
NOF is selected in 14.4\% of annotations.
Each annotation contains an average of 1.78 selected
labels, reflecting the multi-label nature of
the responses. 
To complement these raw counts, the Figure~\ref{fig:affective_overview}(d)
reports Jaccard-normalized co-occurrence among
the eight Mikels-8 categories.
For each pair, the Jaccard index measures the
proportion of annotations containing both categories
among those containing at least one of them.
Amusement and contentment show the strongest
normalized co-occurrence (0.492), followed by
fear and disgust (0.309) and amusement and
excitement (0.304).

\paragraph{Annotation Reliability.}
Finally, we assess inter-annotator agreement
for VAD ratings and categorical annotations.
We additionally examine the reliability
of mean ratings aggregated across the
annotations collected for each panorama.

\textit{VAD ratings.}
\begin{table}[t]
\centering
\caption{
Variance decomposition and model-based reliability
estimates for VAD annotations.
}
\label{tab:annotation_reliability}
\small
\setlength{\tabcolsep}{8pt}
\renewcommand{\arraystretch}{1.15}
\resizebox{\columnwidth}{!}{%
\begin{tabular}{@{}lccc@{}}
\toprule
\textbf{Metric}
& \textbf{Valence}
& \textbf{Arousal}
& \textbf{Dominance} \\
\midrule
Single-rater ICC
& 0.365 & 0.117 & 0.172 \\
Annotator mean reliability
& 0.882 & 0.632 & 0.730 \\
Participant variance share
& 0.055 & 0.127 & 0.210 \\
Residual variance share
& 0.581 & 0.756 & 0.618 \\

\bottomrule
\end{tabular}%
}
\end{table}
For each VAD dimension, we fit a crossed
random-effects model with panorama and participant
as random effects.
From the estimated variance components, we compute
the absolute-agreement single-rater intraclass
correlation coefficient (ICC), which quantifies
the proportion of total rating variance attributable
to differences between panoramas.
We also compute the average-measures ICC,
which estimates the reliability of the mean
rating across annotators for each panorama.
Participant and residual variance shares are
reported to characterize the remaining variation.
As shown in Table~\ref{tab:annotation_reliability},
single-rater reliability is limited across
all three dimensions.
The average-measures ICCs are higher,
at 0.882 for valence, 0.632 for arousal,
and 0.730 for dominance.
The variance decomposition also reveals
different patterns across dimensions:
arousal has the largest residual variance share
(75.6\%), whereas dominance has the largest
participant variance share (21.0\%).
The latter indicates a greater relative
contribution of systematic participant differences
to dominance ratings.

\begin{table}[t]
\centering
\caption{
Inter-annotator agreement for categorical emotion
annotations, measured by binary Krippendorff's $\alpha$.
Higher values indicate greater agreement beyond chance.
}
\label{tab:emotion_alpha}
\small
\setlength{\tabcolsep}{5pt}
\renewcommand{\arraystretch}{1.15}

\begin{tabular*}{\columnwidth}{
@{\extracolsep{\fill}}lrlr@{}
}
\toprule
\textbf{Emotion} & $\boldsymbol{\alpha}$
& \textbf{Emotion} & $\boldsymbol{\alpha}$ \\
\midrule
Amusement   & 0.250 & Sadness & 0.079 \\
Contentment & 0.271 & Disgust & 0.191 \\
Awe         & 0.146 & Anger   & 0.077 \\
Excitement  & 0.162 & Fear    & 0.252 \\
\midrule
\multicolumn{3}{@{}l}{No obvious feeling (NOF)}
& 0.075 \\
\bottomrule
\end{tabular*}
\end{table}

\textit{Categorical annotations.}
Since each annotation can contain multiple emotion
labels, we assess agreement separately for each
Mikels-8 category and the NOF option.
For a given label, we encode each valid annotation
as 1 if the participant selected that label to
describe their emotional response to the scene,
and 0 otherwise.
We compute nominal Krippendorff's $\alpha$
from these binary judgments to assess
inter-annotator agreement.
As shown in Table~\ref{tab:emotion_alpha},
agreement beyond chance is limited,
with $\alpha$ ranging from 0.077 to 0.271
across the eight Mikels-8 categories.
Contentment, fear, and amusement show the
highest agreement, whereas anger and sadness
show the lowest.
The NOF option also exhibits low agreement
(0.075).

\section{Experiments}
We evaluate existing visual models on the task of predicting human emotional responses to 3D environments,
which involves estimating how people feel when experiencing a given 3D space.

\paragraph{Input Representations.}
We consider two image-based representations of
3D spaces as model inputs for predicting human
emotional responses: the \textit{equirectangular panorama} (ERP)
which encodes a 360-degree view of the surrounding space in a single image, and the \textit{cubemap}, representing the surrounding space using six square images corresponding to the faces of a cube (i.e., the front, left, right, back, up, and down directions from a common viewing position).

\paragraph{Prediction Targets.}
We predict both dimensional and categorical
emotional responses to 3D spaces.
For dimensional emotions, the targets are the mean valence, arousal, and dominance
ratings of our collected annotations.
For categorical emotions, we predict a separate Bernoulli distribution for each Mikels-8 category
and the NOF option,
representing the likelihood of a participant would select
the corresponding label (1) or not (0).
The target probability for each label is
estimated as the proportion of annotators
who selected it.
These label-wise probabilities lie in $[0,1]$
but are not constrained to sum to one,
as multiple emotion labels can co-occur.

\begin{table*}[t]
\centering
\caption{
Main results on EmoRama3D with different input
representations and visual models.
}
\label{tab:main_results}
\scriptsize
\setlength{\tabcolsep}{5pt}
\renewcommand{\arraystretch}{1.35}
\setlength{\arrayrulewidth}{0.4pt}

\resizebox{\textwidth}{!}{%
\begin{tabular}{@{}l|ccc|cc|ccc@{}}
\hline

\textbf{Method}
& \textbf{Valence}
& \textbf{Arousal}
& \textbf{Dominance}
& \multicolumn{2}{c|}{\textbf{VAD (Overall)}}
& \multicolumn{3}{c}{\textbf{Discrete Emotion}} \\

\cline{2-4}
\cline{5-6}
\cline{7-9}

& MAE $\downarrow$
& MAE $\downarrow$
& MAE $\downarrow$
& MAE $\downarrow$
& $R^2$ $\uparrow$
& Bernoulli KL $\downarrow$
& MAE $\downarrow$
& Top-3 Recall $\uparrow$ \\

\hline

Train Prior
& 0.753
& 0.416
& 0.600
& 0.590
& -0.002
& 0.123
& 0.148
& 0.513 \\

\hline
\multicolumn{9}{@{}l}{
\textit{Input representation: Cubemap}} \\

SigLIP2 (Frozen backbone)
& 0.601 $\pm$ 0.020
& \textbf{0.379} $\pm$ 0.010
& 0.467 $\pm$ 0.008
& 0.482 $\pm$ 0.012
& 0.307 $\pm$ 0.026
& \textbf{0.086} $\pm$ 0.003
& \textbf{0.112} $\pm$ 0.002
& \textbf{0.670} $\pm$ 0.008 \\

Qwen3-VL (Frozen backbone)
& 0.589 $\pm$ 0.019
& 0.405 $\pm$ 0.010
& 0.460 $\pm$ 0.033
& 0.485 $\pm$ 0.017
& 0.287 $\pm$ 0.061
& \textbf{0.086} $\pm$ 0.008
& 0.112 $\pm$ 0.006
& 0.660 $\pm$ 0.030 \\

Qwen3-VL (QLoRA)
& \textbf{0.571} $\pm$ 0.019
& 0.388 $\pm$ 0.015
& \textbf{0.457} $\pm$ 0.024
& \textbf{0.472} $\pm$ 0.010
& \textbf{0.323} $\pm$ 0.022
& 0.087 $\pm$ 0.003
& 0.112 $\pm$ 0.003
& 0.668 $\pm$ 0.013 \\

Qwen3-VL (Zero-shot)
& 0.656
& 0.629
& 0.675
& 0.653
& -0.387
& 2.654
& 0.199
& 0.360 \\

\hline
\multicolumn{9}{@{}l}{
\textit{Input representation: ERP}} \\

SigLIP2 (Frozen backbone)
& 0.686 $\pm$ 0.005
& 0.407 $\pm$ 0.003
& 0.502 $\pm$ 0.008
& 0.532 $\pm$ 0.005
& 0.178 $\pm$ 0.007
& 0.109 $\pm$ 0.001
& 0.130 $\pm$ 0.001
& 0.611 $\pm$ 0.011 \\

Qwen3-VL (QLoRA)
& \textbf{0.586} $\pm$ 0.009
& \textbf{0.383} $\pm$ 0.009
& \textbf{0.468} $\pm$ 0.020
& \textbf{0.479} $\pm$ 0.003
& \textbf{0.302} $\pm$ 0.011
& \textbf{0.090} $\pm$ 0.003
& \textbf{0.114} $\pm$ 0.003
& \textbf{0.639} $\pm$ 0.013 \\

\hline
\end{tabular}%
}
\vspace{-0.3cm}
\end{table*}

\subsection{Main Evaluation}
\label{sec:experimental_setup}

We evaluate SigLIP2~\cite{tschannen2025siglip2} (SigLIP2-Base-Patch16-224)
and Qwen3-VL~\cite{bai2025qwen3vl} (Qwen3-VL-8B-Instruct)
with frozen backbones, training only
the aggregation and prediction layers.
For Qwen3-VL, we additionally evaluate
QLoRA~\cite{dettmers2023qlora}, which trains
adapters in the top eight language layers,
and zero-shot inference without
task-specific training.
We also include a Train Prior baseline, which predicts deterministic values, i.e., the mean VAD ratings and the selection rates for Mikels-8 and NOF categories from the training dataset.

\paragraph{Implementation Details.}
We use the following composite objective to train
models for both dimensional and categorical emotion prediction:
\begin{equation}
\begin{aligned}
\mathcal{L}
={}&
\mathcal{L}_{\mathrm{mean}}
+\lambda_{\mathrm{var}}\mathcal{L}_{\mathrm{var}}
+\mathcal{L}_{\mathrm{emo}}
+\lambda_{\mathrm{aux}}\mathcal{L}_{\mathrm{aux}} \\
&+
\lambda_{\mathrm{dist}}\mathcal{L}_{\mathrm{dist}}
+\lambda_{\mathrm{top3}}\mathcal{L}_{\mathrm{top3}},
\end{aligned}
\label{eq:training_objective}
\end{equation}
where $\mathcal{L}_{\mathrm{mean}}$ and
$\mathcal{L}_{\mathrm{var}}$ are range-normalized
mean squared error (MSE) losses for predicting
VAD means and variances.
$\mathcal{L}_{\mathrm{emo}}$ and
$\mathcal{L}_{\mathrm{aux}}$ are binary
cross-entropy losses for predicting Mikels-8 emotion
and NOF probabilities, respectively.
$\mathcal{L}_{\mathrm{dist}}$ is a soft-target
cross-entropy loss that matches predicted
VAD distributions to the empirical seven-level
rating histograms.
$\mathcal{L}_{\mathrm{top3}}$ is a pairwise
softplus ranking loss over the nine labels
(Mikels-8 and NOF), encouraging labels in
the ground-truth top-three set, including
ties, to rank above the remaining labels.
We set
$\lambda_{\mathrm{var}}
=\lambda_{\mathrm{aux}}=0.25$
and
$\lambda_{\mathrm{dist}}
=\lambda_{\mathrm{top3}}=0.1$
for all trainable configurations. Unless otherwise specified, models receive
six cubemap images at $512\times512$ resolution. ERP inputs used in the ablations have
a resolution of $1024\times512$.
Models are trained with the AdamW optimizer on 4 NVIDIA H200 GPUs for the main experiments and 8 for the data augmentation experiments. Early stopping is used only in the main experiments.

\paragraph{Metrics and Evaluation Protocol.}
We evaluate VAD predictions using mean absolute error (MAE) for each dimension and report overall MAE and $R^2$ by averaging across all dimensions.
For categorical emotions, we measure the accuracy of predicted label probabilities using MAE and Bernoulli KL divergence, averaged across
the eight emotion categories and NOF.
We also report Top-3 Recall
to evaluate the predicted emotion rankings.

We use 5-fold cross-validation to evaluate model generalization to unseen 3D environments, keeping all panoramas from the same scene within the same split. In each round, one fold is used for testing, and 12.5\% of the remaining scenes are reserved for validation, with the rest used for training.
We train each model with
three random seeds (42, 677, and 3407),
using the same data splits across seeds.
For each seed, we combine the test predictions
from all five folds and calculate the metrics
over the full set of predictions.
We report the mean and sample standard deviation
of these metrics across the three seeds.
Train Prior is deterministic, and the zero-shot
model is evaluated once on the full dataset. Both are reported without standard deviations.

\paragraph{Results.}
\label{sec:main_results}
Table~\ref{tab:main_results} compares model performance
with cubemap and ERP inputs.
With cubemap inputs, all three supervised models
outperform Train Prior in both VAD and categorical
emotion prediction.
Frozen SigLIP2 and Qwen3-VL achieve these improvements
by training only the aggregation and prediction layers.
In contrast, zero-shot Qwen3-VL performs worse than
Train Prior on the overall VAD and categorical
emotion metrics.
For VAD prediction, QLoRA improves Qwen3-VL's
overall MAE from 0.485 to 0.472 and $R^2$
from 0.287 to 0.323, achieving the best overall
results.
It also yields the lowest errors for valence
and dominance, while frozen SigLIP2 performs
best on arousal.
The three supervised models are more closely
matched on categorical emotion prediction,
with MAEs of approximately 0.112 and Bernoulli
KL divergences of 0.086--0.087.
Frozen SigLIP2 achieves the highest Top-3 Recall
at 0.670, compared with 0.668 for QLoRA.
Overall, supervised adaptation is essential. Even lightweight tuning of frozen backbones surpasses both the Train Prior and zero-shot visual model inference.

Comparing input representations, cubemap yields
lower overall VAD and categorical emotion errors
than ERP for both SigLIP2 (Frozen backbone) and
Qwen3-VL (QLoRA).
The difference is larger for SigLIP2:
overall VAD MAE decreases from 0.532 to 0.482,
and categorical emotion MAE from 0.130 to 0.112.
For Qwen3-VL (QLoRA), the corresponding changes
are smaller, from 0.479 to 0.472 and
from 0.114 to 0.112.
Cubemap also improves Top-3 Recall for both
models, from 0.611 to 0.670 for SigLIP2
and from 0.639 to 0.668 for QLoRA.
The advantage does not extend to every dimension:
QLoRA has a slightly lower arousal MAE with ERP
than with cubemap inputs (0.383 vs.\ 0.388).
These results indicate that cubemap representations provide a consistent overall benefit over ERP.

\begin{table}[t]
\centering
\caption{
Effect of data augmentation with ERP inputs.
}
\label{tab:yaw_ablation}
\small
\setlength{\tabcolsep}{5pt}
\renewcommand{\arraystretch}{1.35}
\setlength{\arrayrulewidth}{0.4pt}

\resizebox{\columnwidth}{!}{%
\begin{tabular}{@{}l|c|cc@{}}
\hline

\textbf{Setting}
& \multicolumn{1}{c|}{\textbf{VAD (Overall)}}
& \multicolumn{2}{c}{\textbf{Discrete Emotion}} \\

\cline{2-4}

& MAE $\downarrow$
& MAE $\downarrow$
& Top-3 Recall $\uparrow$ \\

\hline

\multicolumn{4}{@{}l}{
\textit{SigLIP2 (Frozen backbone)}} \\

Original
& 0.531 $\pm$ 0.005
& 0.131 $\pm$ 0.001
& 0.614 $\pm$ 0.014 \\

Repeat $\times10$
& 0.517 $\pm$ 0.007
& 0.126 $\pm$ 0.002
& \textbf{0.620} $\pm$ 0.004 \\

Yaw $\times10$
& \textbf{0.505} $\pm$ 0.014
& \textbf{0.124} $\pm$ 0.002
& 0.618 $\pm$ 0.013 \\

\hline

\multicolumn{4}{@{}l}{
\textit{Qwen3-VL (QLoRA)}} \\

Original
& 0.481 $\pm$ 0.005
& 0.114 $\pm$ 0.003
& 0.644 $\pm$ 0.015 \\

Repeat $\times10$
& 0.500 $\pm$ 0.024
& 0.121 $\pm$ 0.005
& 0.632 $\pm$ 0.019 \\

Yaw $\times10$
& \textbf{0.465} $\pm$ 0.002
& \textbf{0.112} $\pm$ 0.000
& \textbf{0.661} $\pm$ 0.009 \\

\hline
\end{tabular}%
}
\end{table}
\subsection{Effect of Data Augmentation}
\label{sec:data_augmentation}

We examine whether yaw-based data augmentation
improves emotion prediction from ERP inputs.
For each training panorama, we generate ten
versions with yaw angles ranging from
$0^\circ$ to $324^\circ$ at $36^\circ$ intervals.
These versions retain the original annotations
while varying the horizontal orientation
of the input.
We compare three training settings:
\textit{Original}, which uses each panorama once;
\textit{Repeat $\times10$}, which repeats
the original panorama ten times; and
\textit{Yaw $\times10$}, which uses the ten
orientations described above.
The repetition setting allows us to examine
whether augmentation offers benefits beyond
additional exposure to the same training data.
We keep the optimization epochs fixed for different settings.

As shown in Table~\ref{tab:yaw_ablation},
Yaw $\times10$ achieves the lowest VAD and
discrete emotion MAEs for both models.
Compared with Repeat $\times10$, it reduces
VAD MAE from 0.517 to 0.505 for SigLIP2
and from 0.500 to 0.465 for Qwen3-VL (QLoRA).
Emotion MAE also decreases from 0.126 to 0.124
and from 0.121 to 0.112, respectively.
For Top-3 Recall, Yaw $\times10$ improves
Qwen3-VL from 0.632 to 0.661, while
SigLIP2 performs slightly better with repetition
(0.620 vs.\ 0.618).
Overall, these results suggest that data augmentation through varying
the horizontal viewing orientation benefits
emotion prediction beyond simply repeating
the original training images.
\section{Discussion}
\label{sec:discussion}

The experiments show that pretrained visual
representations contain useful information
for predicting human emotional responses to 3D spaces.
However, model adaptation does not improve performance uniformly across all dimensional and categorical prediction tasks.
These findings suggest that dimensional
and categorical emotion prediction present
different challenges.

\paragraph{Potential Applications.}
Beyond emotion prediction, EmoRama3D offers
opportunities for affect-aware spatial design.
Unlike datasets containing only fixed images,
EmoRama3D provides editable 3D scene assets,
allowing researchers to modify environmental
properties such as lighting, materials,
and object arrangements.
These capabilities could support controlled
studies of how spatial design choices
influence human emotional responses.
Furthermore, emotion prediction models
trained on EmoRama3D could potentially
provide feedback during spatial design,
helping designers compare alternatives
and explore environments associated with
desired emotional responses.
Such applications would require further
validation to establish whether model
predictions correspond to actual changes
in human emotional experience.

\paragraph{Limitations and Future Work.} 
Despite its contributions to understanding human
emotional responses to 3D environments,
EmoRama3D has several limitations.
First, we use panoramic images to represent
3D environments for annotation,
as real-time rendering and interaction with
complex 3D scenes in VR can be computationally
demanding.
Although panoramas provide immersive visual
experiences, they restrict participants to
fixed viewpoints and do not support spatial
navigation or interaction.
Future work could explore real-time 3D
environments to capture emotional responses
during more naturalistic spatial experiences.
Second, emotional responses to environments
may depend on their semantic and contextual
meanings.
Our current study focuses on visually induced
emotional responses without explicitly
considering semantic context, such as the
function, familiarity, or cultural significance
of a space.
Incorporating such contextual information
could provide a more comprehensive understanding
of spatial emotion perception.
Finally, although EmoRama3D provides editable
3D scene assets, their physical plausibility
is not fully guaranteed.
For example, modifications to architectural
structures may violate structural stability
or other physical constraints.
Future work could incorporate physical
and architectural constraints to support
more realistic and reliable scene modifications
for affect-aware spatial design.

\section{Conclusion}
\label{sec:conclusion}

We introduced EmoRama3D, a dataset for studying
human emotional responses to 3D environments.
It combines editable 3D scenes with annotations
collected through immersive VR, covering
VAD ratings and multiple emotion categories.
We also evaluated existing visual models
on predicting these responses from image-based
representations of 3D spaces.
The results show that pretrained visual
representations can support the prediction
of emotional responses to 3D spaces, although
both dimensional and categorical predictions
leave room for improvement.
By retaining editable source scenes and
individual annotations, EmoRama3D provides
a foundation for further research on emotion
prediction
and affect-aware spatial design. We plan to publicly release EmoRama3D, including
the 3D scene assets, human emotion annotations,
and code, to facilitate future research
on affective understanding and design of 3D environments.

{
\small
\bibliographystyle{ieeenat_fullname}
\bibliography{main}
}

\clearpage
\appendix
\section{Annotation Details}
\label{sec:human_annotation_details}

We recruited 22 participants through a recruitment post published on a social media platform and all annotations were conducted in person.
For privacy protection, we assigned anonymous IDs to all participants (P001–P022). We divided the panoramas to be annotated into three parts containing 63, 62, and 61 panoramas, respectively. Participants could voluntarily choose to annotate one, two, or all three parts according to their availability and willingness. Participant assignments across the three parts are summarized in Table~\ref{tab:participant_details}.
Consequently, each panorama received 13 independent annotations, resulting in 2,418 annotation instances in total.
Participants received \$0.10 for each completed panorama annotation, together with a fixed \$2.80 travel allowance to compensate for in-person attendance. Given that annotating each part (approximately 61–63 panoramas) took about one hour, the resulting compensation exceeded the applicable local minimum wage at the time of the study.

\begin{table}[ht]
\centering
\caption{Demographic information and annotation assignments
of all 22 annotators. Participant IDs are anonymized.}
\label{tab:participant_details}
\setlength{\tabcolsep}{3.5pt}
\renewcommand{\arraystretch}{1.12}
\begin{tabular}{cccc}
\hline
\textbf{ID} &
\textbf{Age Range} &
\textbf{Gender} &
\textbf{Annotation Parts} \\
\hline
P001 & 30--39 & Female & 3 \\
P002 & 20--29 & Female & 1  \\
P003 & 20--29 & Female & 2, 3  \\
P004 & 30--39 & Male & 1  \\
P005 & 20--29 & Male & 2, 3 \\
P006 & 30--39 & Female & 1 \\
P007 & 30--39 & Male & 1, 2, 3 \\
P008 & 20--29 & Female & 3  \\
P009 & 20--29 & Female & 1, 2 \\
P010 & 20--29 & Male & 2 \\
P011 & 20--29 & Female & 1, 3  \\
P012 & 20--29 & Male & 2, 3 \\
P013 & 20--29 & Female & 1, 2 \\
P014 & 20--29 & Male & 3 \\
P015 & 20--29 & Female & 1, 2 \\
P016 & 30--39 & Female & 3  \\
P017 & 20--29 & Female & 1 \\
P018 & 50--59 & Female & 2 \\
P019 & 20--29 & Male & 1, 2, 3  \\
P020 & 20--29 & Male & 1, 2, 3  \\
P021 & 20--29 & Female & 1, 2, 3 \\
P022 & 20--29 & Male & 1, 2, 3 \\
\hline
\end{tabular}
\end{table}

\section{Dataset Details}
\label{sec:dataset_resources_licensing}

We collected candidate 3D scenes from publicly accessible platforms. We first identified around 200 candidate scenes. Each candidate was assessed for its suitability for panoramic rendering. Assets with missing textures, incomplete geometry, import or rendering failures, unsuitable camera configurations, low visual quality, or substantial content redundancy were not selected. After this screening, 159 scenes were retained for EmoRama3D.
The final dataset comprises 159 3D scenes. The assets include 135 Blender files (\texttt{.blend}) and 24 glTF binary files (\texttt{.glb}). We rendered 186 panoramas from the 159 scenes. Specifically, 141 scenes contribute one panorama, 11 contribute two, 5 contribute three, and 2 contribute four panoramas.

\begin{figure*}[ht]
    \centering
    \includegraphics[width=\linewidth]{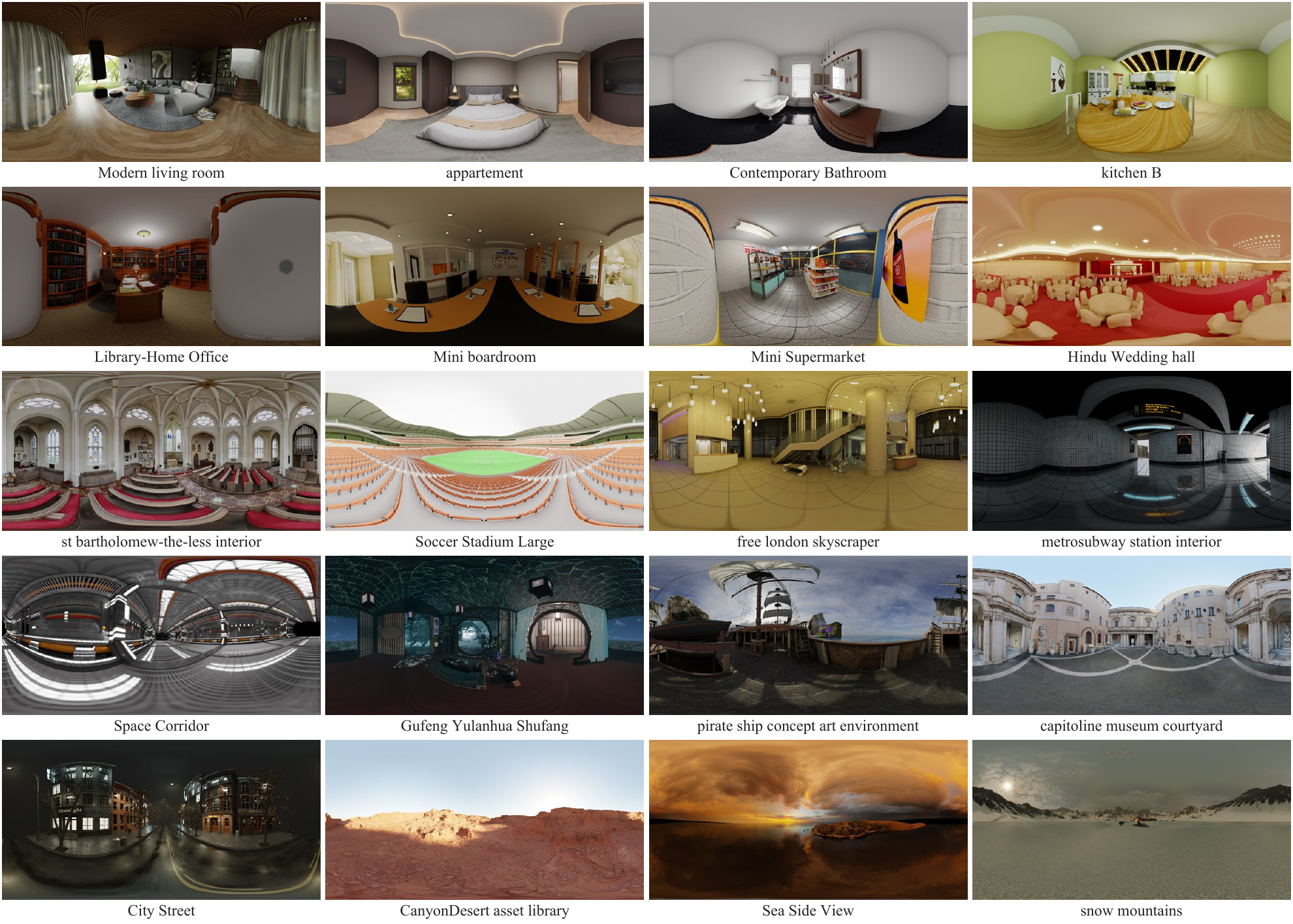}
    \caption{Representative examples of diverse 3D scenes in EmoRama3D.}
    \label{fig:scene_showcase}
    \vspace{-0.3cm}
\end{figure*}

\begin{table}[ht]
\centering
\caption{License distribution. ``Other license'' denotes non-Creative Commons licenses that permit commercial use and modification.}
\label{tab:source_summary}
\small
\setlength{\tabcolsep}{3.5pt}
\renewcommand{\arraystretch}{1.12}
\begin{tabular}{@{}lrr@{}}
\hline
\textbf{License} & \textbf{Scenes} & \textbf{Panoramas} \\
\hline
CC0                         & 53 & 63 \\
CC-BY                       & 74 & 80 \\
CC-BY 4.0                   & 22 & 31 \\
CC-BY-SA                    & 5  & 5  \\
CC-BY-NC                    & 1  & 1  \\
CC-BY-NC-SA                 & 1  & 1  \\
Other license               & 3  & 5  \\
\hline
\textbf{Total}              & \textbf{159} & \textbf{186} \\
\hline
\end{tabular}
\end{table}

Table~\ref{tab:source_summary} summarizes the license distribution of the scene entries used to produce panoramic renderings. Counts are reported separately for scene entries and panoramic renderings, as some scenes were rendered from multiple viewpoints. The retained collection includes CC0 (53 scenes), CC-BY (74), CC-BY 4.0 (22), CC-BY-SA (5), CC-BY-NC (1), CC-BY-NC-SA (1), and other licenses (3) that permit commercial use and modification but are not classified as Creative Commons licenses.
We will release the dataset in accordance with the corresponding source-asset licenses and will provide attribution information where required.

Figure~\ref{fig:scene_showcase} shows diverse panoramas from EmoRama3D, including indoor spaces, public facilities, and natural landscapes, to illustrate the diversity of scenes in our dataset.

\section{Additional Results}
We provide additional results from ablation studies on the loss function terms and on data augmentation on cubemap.
\subsection{Loss Function Ablation}
\label{sec:sup_loss}

We ablate the overall training objective as described in the main paper:
\begin{equation}
\begin{aligned}
\mathcal{L} =\;&
\mathcal{L}_{\mathrm{mean}}
+\lambda_{\mathrm{var}}\mathcal{L}_{\mathrm{var}}
+\mathcal{L}_{\mathrm{emo}}
+\lambda_{\mathrm{aux}}\mathcal{L}_{\mathrm{aux}}\\
&+\lambda_{\mathrm{dist}}\mathcal{L}_{\mathrm{dist}}
+\lambda_{\mathrm{top3}}\mathcal{L}_{\mathrm{top3}}.
\end{aligned}
\label{eq:supp_training_objective}
\end{equation}
We examine the effect of the two optional supervision terms: the VAD-distribution loss \(\mathcal{L}_{\mathrm{dist}}\) and the top-three ranking loss \(\mathcal{L}_{\mathrm{top3}}\). We compare four configurations: the full objective, which includes both terms; \(w/o\ \mathcal{L}_{\mathrm{top3}}\), which removes only the ranking term; \(w/o\ \mathcal{L}_{\mathrm{dist}}\), which removes only the distribution term; and \(w/o\ \mathcal{L}_{\mathrm{top3}}+\mathcal{L}_{\mathrm{dist}}\), which removes both optional terms. All other training objectives and settings are kept unchanged.

\begin{table*}[t]
\centering
\caption{
Ablation study on loss function with six-view cubemap inputs.
}
\label{tab:sup_loss_ablation}
\scriptsize
\setlength{\tabcolsep}{5pt}
\renewcommand{\arraystretch}{1.35}
\setlength{\arrayrulewidth}{0.4pt}

\resizebox{\textwidth}{!}{%
\begin{tabular}{@{}l|ccc|cc|ccc@{}}
\hline

\textbf{Setting}
& \textbf{Valence}
& \textbf{Arousal}
& \textbf{Dominance}
& \multicolumn{2}{c|}{\textbf{VAD (Overall)}}
& \multicolumn{3}{c}{\textbf{Discrete Emotion}} \\

\cline{2-4}
\cline{5-6}
\cline{7-9}

& MAE \(\downarrow\)
& MAE \(\downarrow\)
& MAE \(\downarrow\)
& MAE \(\downarrow\)
& \(R^2\) \(\uparrow\)
& Bernoulli KL \(\downarrow\)
& MAE \(\downarrow\)
& Top-3 Recall \(\uparrow\) \\

\hline
\multicolumn{9}{@{}l}{
\textit{SigLIP2 (Frozen backbone)}} \\

\(w/o\ \mathcal{L}_{\mathrm{top3}}+\mathcal{L}_{\mathrm{dist}} \)
& 0.603 \(\pm\) 0.027
& \textbf{0.377} \(\pm\) 0.006
& \textbf{0.463} \(\pm\) 0.006
& \textbf{0.481} \(\pm\) 0.009
& \textbf{0.311} \(\pm\) 0.016
& 0.084 \(\pm\) 0.003
& 0.112 \(\pm\) 0.003
& \textbf{0.677} \(\pm\) 0.015 \\

\(w/o\ \mathcal{L}_{\mathrm{top3}}\)
& 0.606 \(\pm\) 0.022
& 0.384 \(\pm\) 0.002
& 0.469 \(\pm\) 0.017
& 0.486 \(\pm\) 0.012
& 0.300 \(\pm\) 0.017
& 0.084 \(\pm\) 0.002
& 0.112 \(\pm\) 0.001
& 0.674 \(\pm\) 0.004 \\

\(w/o\ \mathcal{L}_{\mathrm{dist}}\)
& \textbf{0.599} \(\pm\) 0.015
& 0.384 \(\pm\) 0.007
& 0.465 \(\pm\) 0.005
& 0.482 \(\pm\) 0.007
& 0.308 \(\pm\) 0.016
& \textbf{0.083} \(\pm\) 0.001
& \textbf{0.110} \(\pm\) 0.002
& 0.671 \(\pm\) 0.001 \\

Full objective
& 0.601 \(\pm\) 0.020
& 0.379 \(\pm\) 0.010
& 0.467 \(\pm\) 0.008
& 0.482 \(\pm\) 0.012
& 0.307 \(\pm\) 0.026
& 0.086 \(\pm\) 0.003
& 0.112 \(\pm\) 0.002
& 0.670 \(\pm\) 0.008 \\

\hline
\multicolumn{9}{@{}l}{
\textit{Qwen3-VL (QLoRA)}} \\

\(w/o\ \mathcal{L}_{\mathrm{top3}}+\mathcal{L}_{\mathrm{dist}}\)
& 0.584 \(\pm\) 0.050
& 0.387 \(\pm\) 0.012
& 0.445 \(\pm\) 0.003
& 0.472 \(\pm\) 0.020
& 0.326 \(\pm\) 0.045
& 0.087 \(\pm\) 0.008
& 0.114 \(\pm\) 0.007
& 0.661 \(\pm\) 0.033 \\

\(w/o\ \mathcal{L}_{\mathrm{top3}}\)
& \textbf{0.564} \(\pm\) 0.010
& 0.387 \(\pm\) 0.019
& 0.462 \(\pm\) 0.007
& 0.471 \(\pm\) 0.010
& 0.331 \(\pm\) 0.028
& \textbf{0.083} \(\pm\) 0.003
& \textbf{0.111} \(\pm\) 0.003
& 0.666 \(\pm\) 0.008 \\

\(w/o\ \mathcal{L}_{\mathrm{dist}}\)
& 0.583 \(\pm\) 0.040
& \textbf{0.386} \(\pm\) 0.012
& \textbf{0.436} \(\pm\) 0.017
& \textbf{0.468} \(\pm\) 0.005
& \textbf{0.339} \(\pm\) 0.010
& 0.086 \(\pm\) 0.007
& 0.113 \(\pm\) 0.007
& 0.661 \(\pm\) 0.023 \\

Full objective
& 0.571 \(\pm\) 0.019
& 0.388 \(\pm\) 0.015
& 0.457 \(\pm\) 0.024
& 0.472 \(\pm\) 0.010
& 0.323 \(\pm\) 0.022
& 0.087 \(\pm\) 0.003
& 0.112 \(\pm\) 0.003
& \textbf{0.668} \(\pm\) 0.013 \\

\hline
\end{tabular}%
}
\end{table*}

\begin{table*}[!t]
\centering
\caption{
Results of data augmentation on cubemap inputs.
}
\label{tab:sup_yaw_ablation}
\scriptsize
\setlength{\tabcolsep}{5pt}
\renewcommand{\arraystretch}{1.35}
\setlength{\arrayrulewidth}{0.4pt}

\resizebox{\textwidth}{!}{%
\begin{tabular}{@{}l|ccc|cc|ccc@{}}
\hline

\textbf{Setting}
& \textbf{Valence}
& \textbf{Arousal}
& \textbf{Dominance}
& \multicolumn{2}{c|}{\textbf{VAD (Overall)}}
& \multicolumn{3}{c}{\textbf{Discrete Emotion}} \\

\cline{2-4}
\cline{5-6}
\cline{7-9}

& MAE $\downarrow$
& MAE $\downarrow$
& MAE $\downarrow$
& MAE $\downarrow$
& $R^2$ $\uparrow$
& Bernoulli KL $\downarrow$
& MAE $\downarrow$
& Top-3 Recall $\uparrow$ \\

\hline
\multicolumn{9}{@{}l}{
\textit{SigLIP2 (Frozen backbone)}} \\

Original
& 0.602 $\pm$ 0.039
& 0.394 $\pm$ 0.010
& 0.474 $\pm$ 0.020
& 0.490 $\pm$ 0.016
& 0.279 $\pm$ 0.024
& 0.090 $\pm$ 0.005
& 0.114 $\pm$ 0.002
& 0.658 $\pm$ 0.007 \\

Repeat $\times10$
& \textbf{0.581} $\pm$ 0.014
& \textbf{0.387} $\pm$ 0.015
& 0.441 $\pm$ 0.007
& \textbf{0.470} $\pm$ 0.003
& \textbf{0.334} $\pm$ 0.014
& \textbf{0.083} $\pm$ 0.001
& 0.111 $\pm$ 0.001
& 0.661 $\pm$ 0.007 \\

Yaw $\times10$
& 0.591 $\pm$ 0.011
& \textbf{0.387} $\pm$ 0.012
& \textbf{0.433} $\pm$ 0.006
& 0.471 $\pm$ 0.002
& 0.330 $\pm$ 0.012
& \textbf{0.083} $\pm$ 0.001
& \textbf{0.110} $\pm$ 0.000
& \textbf{0.670} $\pm$ 0.005 \\

\hline
\multicolumn{9}{@{}l}{
\textit{Qwen3-VL (QLoRA)}} \\

Original
& 0.591 $\pm$ 0.023
& \textbf{0.382} $\pm$ 0.008
& \textbf{0.457} $\pm$ 0.052
& \textbf{0.477} $\pm$ 0.021
& \textbf{0.331} $\pm$ 0.035
& \textbf{0.086} $\pm$ 0.001
& 0.112 $\pm$ 0.001
& 0.666 $\pm$ 0.002 \\

Repeat $\times10$
& 0.613 $\pm$ 0.058
& 0.384 $\pm$ 0.023
& 0.468 $\pm$ 0.024
& 0.488 $\pm$ 0.030
& 0.316 $\pm$ 0.069
& 0.105 $\pm$ 0.015
& 0.117 $\pm$ 0.005
& 0.650 $\pm$ 0.019 \\

Yaw $\times10$
& \textbf{0.570} $\pm$ 0.014
& 0.403 $\pm$ 0.006
& 0.463 $\pm$ 0.022
& 0.478 $\pm$ 0.004
& 0.327 $\pm$ 0.018
& 0.090 $\pm$ 0.004
& \textbf{0.112} $\pm$ 0.002
& \textbf{0.674} $\pm$ 0.012 \\

\hline
\end{tabular}%
}
\vspace{-0.3cm}
\end{table*}

Table~\ref{tab:sup_loss_ablation} shows that the four configurations produce only small differences. For SigLIP2, \(w/o\ \mathcal{L}_{\mathrm{top3}}+\mathcal{L}_{\mathrm{dist}}\) achieves the lowest overall VAD MAE (0.481) and the highest Top-3 Recall (0.677), whereas \(w/o\ \mathcal{L}_{\mathrm{dist}}\) yields the lowest discrete emotion MAE (0.110) and Bernoulli KL (0.083). For Qwen3-VL, \(w/o\ \mathcal{L}_{\mathrm{dist}}\) achieves the lowest overall VAD MAE (0.468), \(w/o\ \mathcal{L}_{\mathrm{top3}}\) yields the lowest discrete emotion MAE (0.111) and Bernoulli KL (0.083), and the full objective obtains the highest Top-3 Recall (0.668). Overall VAD MAE ranges only from 0.481 to 0.486 for SigLIP2 and from 0.468 to 0.472 for Qwen3-VL, while discrete emotion MAE varies by at most 0.003 within each backbone. Moreover, adding the ranking term alone does not improve Top-3 Recall over \(w/o\ \mathcal{L}_{\mathrm{top3}}+\mathcal{L}_{\mathrm{dist}}\) (0.677 to 0.671 for SigLIP2; unchanged at 0.661 for Qwen3-VL), and adding the distribution term alone yields inconsistent VAD gains. These results indicate limited and backbone-dependent sensitivity to the two optional terms.

\subsection{Data Augmentation on Cubemap}
\label{sec:sup_yaw}

\begin{figure}
    \centering
    \includegraphics[width=\linewidth]{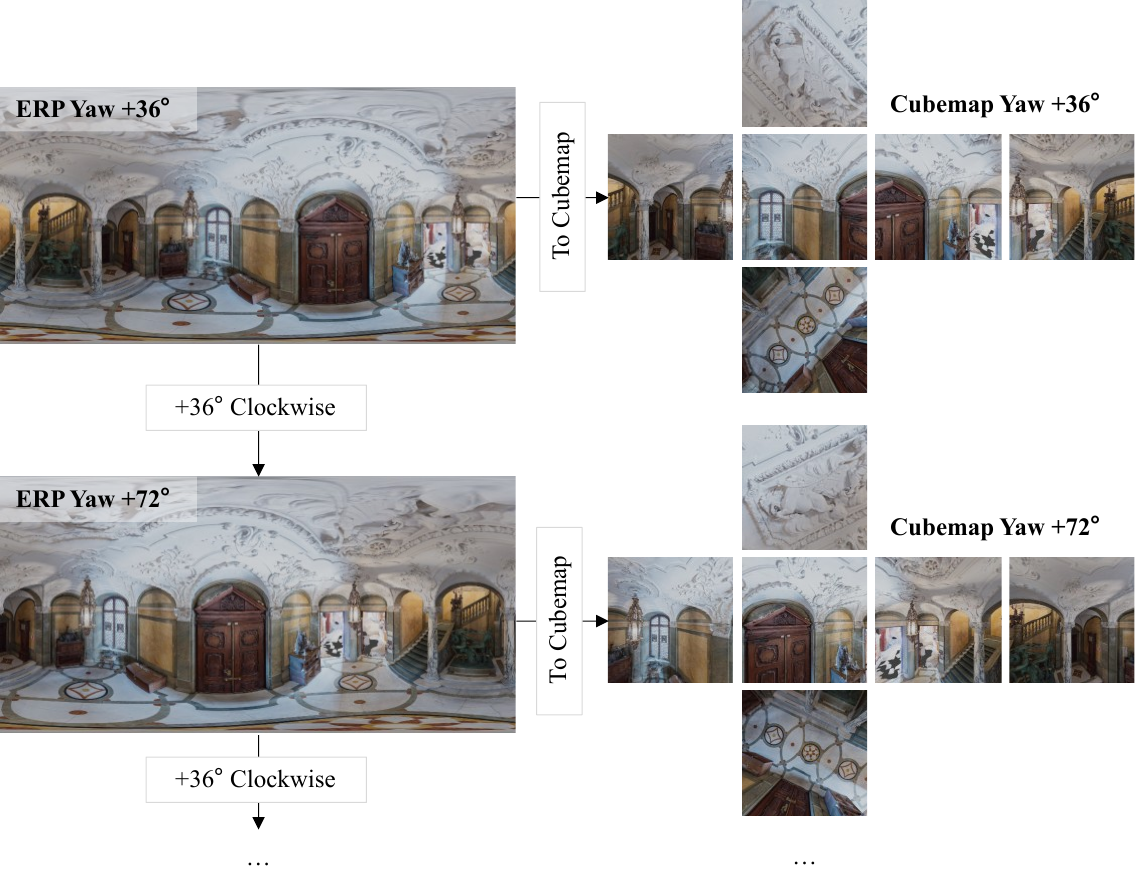}
        \vspace{-0.4cm}
    \caption{Illustration of cubemap yaw augmentation. Two of the ten yaw orientations, spaced at $36^\circ$ intervals, are shown. Each rotated ERP panorama is projected onto six cube faces and used as six separate input views.}
    \label{fig:yaw_schematic}
    \vspace{-0.3cm}
\end{figure}

The main paper studies data augmentation for single-image ERP inputs but not for cubemap inputs. Here we extend the analysis to six-view cubemaps and compare different data augmentation strategies for cubemap inputs.
We compare three settings: Original uses one orientation per panorama; Repeat $\times10$ replicates the original orientation ten times; and Yaw $\times10$ uses ten distinct orientations at $36^\circ$ intervals over $360^\circ$. Figure~\ref{fig:yaw_schematic} illustrates the yaw augmentation pipeline. Repeat $\times10$ and Yaw $\times10$ have matched training-item counts, epoch budgets, and optimizer steps, isolating the effect of orientation diversity.

Table~\ref{tab:sup_yaw_ablation} shows that the effects of data augmentation for cubemap inputs depend on the backbone and evaluation metric. Comparing Yaw $\times10$ with Repeat $\times10$ isolates the effect of varying the yaw orientation under matched training budgets, whereas comparisons with Original also reflect increased training exposure.
For SigLIP2, both settings improve overall VAD prediction over Original: VAD MAE decreases from 0.490 to about 0.470, and $R^2$ increases from 0.279 to about 0.33. Yaw $\times10$ performs similarly to Repeat $\times10$ on VAD, but slightly improves discrete emotion MAE and Top-3 Recall. For Qwen3-VL, Repeat $\times10$ degrades overall VAD MAE, discrete emotion MAE, and Top-3 Recall relative to Original, whereas Yaw $\times10$ brings overall VAD MAE and discrete emotion MAE close to their Original values and achieves the highest Top-3 Recall (0.674). However, Yaw $\times10$ does not outperform Original in overall VAD MAE, $R^2$, or Bernoulli KL.
Overall, using different yaw orientations provides only modest gains over repeating the same orientation. Compared with the ERP-based augmentation in the main paper, the improvement over Original is less pronounced for six-view cubemap inputs, possibly because cubemap inputs already expose the model to multiple viewing directions.

\end{document}